\pdfoutput=1

\documentclass[11pt]{article}

\usepackage[]{EMNLP2023}

\usepackage{times}
\usepackage{latexsym}

\usepackage[T1]{fontenc}

\usepackage[utf8]{inputenc}

\usepackage{microtype}

\usepackage{inconsolata}
\usepackage{multirow}
\usepackage{graphicx}
\usepackage{amsmath}
\usepackage{amsthm}
\usepackage{amssymb}
\usepackage{booktabs}
\usepackage{algorithm}
\usepackage{algorithmic}
\usepackage{subfigure}
\usepackage{color}
\usepackage{tabularray}
\usepackage{arydshln}
\usepackage{enumitem}
\title{RoPDA: Robust Prompt-based Data Augmentation for Low-Resource Named Entity Recognition}

\author{Sihan Song,$^1$ Furao Shen,$^2$ Jian Zhao$^3$ \\[0.6em]
$^1$  Department of Computer Science and Technology, Nanjing University\\
$^2$ School of Artificial Intelligence, Nanjing University\\
$^3$  School of Electronic Science and Engineering, Nanjing University\\
[0.6em]
\texttt{whalesihan@smail.nju.edu.cn, \{frshen,jianzhao\}@nju.edu.cn}
}

\begin{document}
\maketitle
\begin{abstract}
Data augmentation has been widely used in low-resource NER tasks to tackle the problem of data sparsity. However, previous data augmentation methods have the disadvantages of disrupted syntactic  structures, token-label mismatch, and requirement for external knowledge or manual effort.
To address these issues, we propose \textbf{Ro}bust \textbf{P}rompt-based \textbf{D}ata \textbf{A}ugmentation (RoPDA) for low-resource NER. Based on pre-trained language models (PLMs) with continuous prompt, RoPDA performs entity augmentation and context augmentation through five fundamental augmentation operations to generate label-flipping and label-preserving examples.
To optimize the utilization of the augmented samples, we present two techniques: Self-Consistency Filtering and mixup. The former effectively eliminates low-quality samples, while the latter prevents performance degradation arising from the direct utilization of label-flipping samples.
Extensive experiments on three benchmarks from different domains demonstrate that RoPDA significantly improves upon strong baselines, and also outperforms state-of-the-art semi-supervised learning methods when unlabeled data is included.
\end{abstract}

\section{Introduction}
Named Entity Recognition (NER) is a fundamental NLP task which is dedicated to identifying predefined named entities  (\emph{e.g.}, persons, organizations and locations) from texts.
With the rapid development of deep learning in recent years, fine-tuning pre-trained language models (PLMs) such as BERT \cite{devlin-etal-2019-bert} for NER tasks has yielded promising results \cite{zhong-chen-2021-frustratingly,chen2022style}. However, fine-tuning PLMs still necessitates a substantial amount of human annotations. NER models, in particular, require each token to be labeled in a sequence, which is a laborious and time-consuming task in the real world. 
Consequently, NER frequently encounters the challenge of data sparsity, making low-resource NER a pressing priority.

\begin{figure}[t]
    \centering
\includegraphics[width=0.82\linewidth]{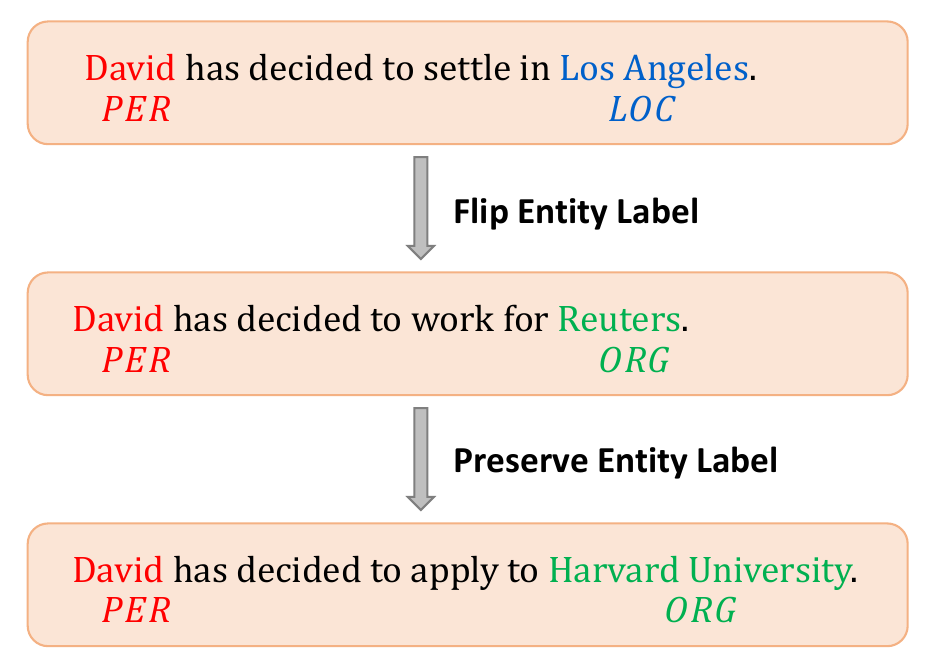}
    \caption{Label-flipping and label-preserving operations.}
    \label{fig:flip_example}
\end{figure}
In order to mitigate the issue of data sparsity, 
various data augmentation methods for low-resource NER have been proposed, such as traditional word-level manipulation \cite{dai2020analysis,zeng2020counterfactual} and more recent PLM-based methods, and the latter have received a lot of attention and have yielded promising results \cite{zhou2022melm,wang2022promda}. \citet{zhou2022melm} leverage a masked language model to randomly mask entities in the sentence and regenerate them conditioned on labels, thereby enhancing entity diversity. \citet{wang2022promda} employ Soft Prompt for seq2seq PLMs and propose a dual-view augmentation approach to generate sentences conditioned on labels and keywords.
However, their approaches have the limitation of only boosting entity diversity but not context diversity, or necessitating external knowledge (\emph{i.e.}, external corpus).

To address these issues, we propose \textbf{Ro}bust \textbf{P}rompt-based \textbf{D}ata \textbf{A}ugmentation (RoPDA) for low-resource NER.  
With well-trained continuous prompt \cite{li2021prefix}, our model is capable of automatically generating training samples for low-resource NER tasks, eliminating the need for external knowledge or human efforts, which is different from \citet{wang2022promda}.
To enhance model generalization,  we propose five fundamental data augmentation operations. Among these, one operation focuses on context augmentation, aiming to increase context diversity and enrich our training data. The remaining four operations concentrate on entity augmentation, designed to generate adversarial examples and promote entity diversity, including label-flipping and label-preserving operations.

Inspired by \citet{zhou-etal-2022-flipda}, a label-flipping operation means regenerating an entity in the sentence into a different type of entity.
As shown in Figure \ref{fig:flip_example}, after a label-flipping operation, the entity type sequences of the augmented sentence and the original sentence differ only in one entity.
Such an augmented sentence can serve as an adversarial example \cite{Goodfellow2014ExplainingAH} for this modified entity type, enhancing the NER model's capability in distinguishing these two entity types before and after regeneration. 
Conversely, A label-preserving operation involves regenerating an entity as a new entity of the same type, thereby prompting entity diversity. 
We argue that NER models can benefit from both label-preserving and label-flipping operations.
With these five augmentation operations, we can generate smooth and diverse sentences without relying on external knowledge.
To further optimize the utilization of augmented samples, we propose Self-Consistency \cite{wang2022self} Filtering and mixup. The former acquires the capability to effectively filter out low-quality samples by fine-tuning the PLM with a bidirectional mask. The latter employs linear interpolation for adversarial examples, resulting in a smoother label distribution that prevents the performance degradation caused by their direct utilization.
Through these five augmentation operations followed by Self-Consistency Filtering and mixup, we can generate high-quality augmented samples.

To summarize, our contributions are as follows:
\begin{itemize}[topsep=0pt,itemsep=-1ex,partopsep=1ex,parsep=1ex]
    \item   We propose a robust prompt-based data augmentation method RoPDA for low-resource NER, which contains five augmentation operations. By simultaneously augmenting entities and contexts, RoPDA can generate augmented examples with high diversity.
\item We propose Self-Consistency Filtering to improve the quality of augmented samples through bidirectional masking.
\item  We utilize the mixup technique to interpolate adversarial examples with corresponding original examples, effectively maximizing the utility of adversarial examples.
\item Experiments on three benchmarks show the significant performance gain of RoPDA over current state-of-the-art baselines.
\end{itemize}

\section{Related Work}
\subsection{Data Augmentation}
Word-level manipulation is a prevalent data augmentation method that manipulates words in the original text to generate synthetic text utilizing predefined rules \cite{dai2020analysis,wei2019eda,kobayashi2018contextual}. \citet{dai2020analysis} generate new examples through token substitutions, including synonym replacement and mention replacement. But these methods run the risk of destroying the sentence structure and making labels inconsistent with modified tokens. 
Recently, leveraging the powerful generative capability and rich knowledge of PLMs to generate augmented data has been explored \cite{kumar2020data,ding2020daga}. \citet{zhou2022melm} perform entity replacement on corrupted sentences based on the masked language model.
\citet{wang2022promda} and \citet{anaby2020not} leverage seq2seq PLMs to generate synthetic data conditioned on entity types. 

Adversarial augmentation \cite{volpi2018generalizing} is commonly used to improve model robustness, but recent research has discovered that it can also improve model generalization \cite{xie2020adversarial}. \citet{reich2022leveraging} adopt expert-guided heuristics for adversarial augmentation, focusing on modifying entities and contexts in accordance with predefined rules.  However, their approach lacks diversity in entities and relies on expert knowledge. \citet{lin-etal-2021-rockner} create entity attacks by replacing the target entity with entities of the same type from an external knowledge base. On the other hand, our approach, which is based on PLMs, can produce high-quality adversarial examples without the need for external knowledge or human effort.
\begin{figure*}[!ht]
  \subfigure[Overview]{
   \centering
    \resizebox{1\linewidth}{!}{
    \includegraphics{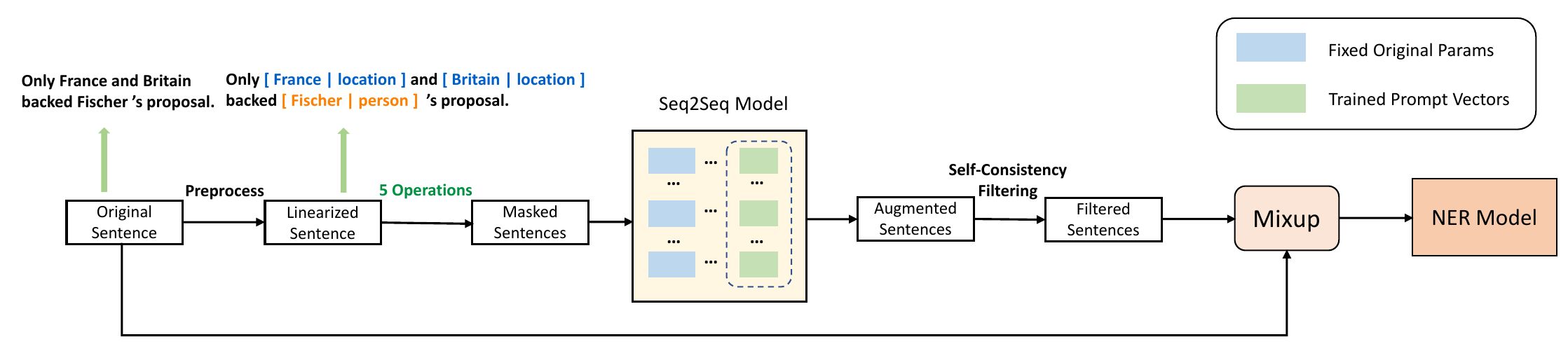}
    }
    \label{fig:overview}
  }
  \subfigure[Masked Sentences]{
    \centering
    \resizebox{0.44\linewidth}{!}{
    \includegraphics{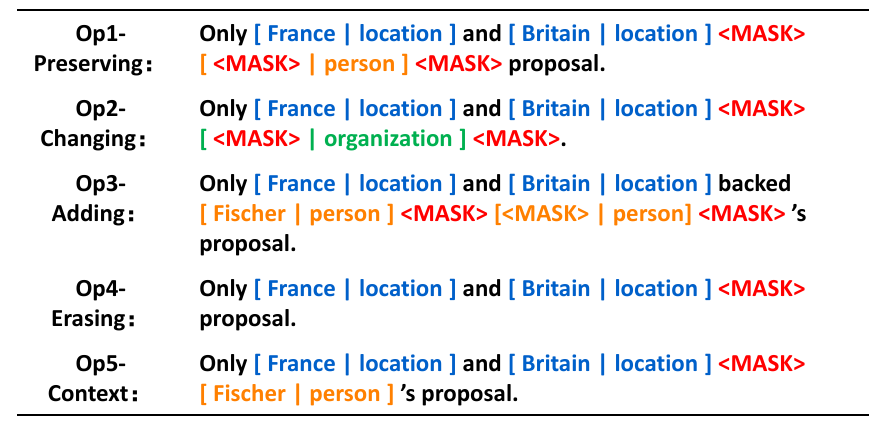}
    }
    \label{fig:overview_ex1}
  }
\subfigure[Augmented Sentences]{
\centering
\resizebox{0.44\linewidth}{!}{
\includegraphics{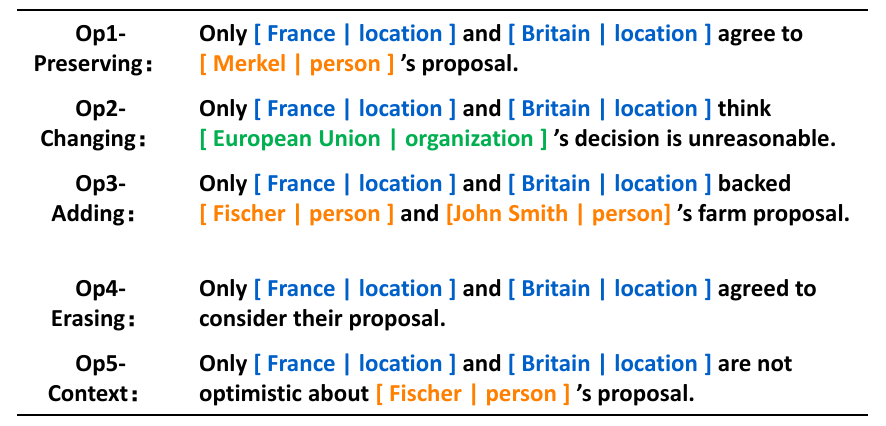}
}
\label{fig:overview_ex2}
}
\caption{Overview of our proposed RoPDA. Figure \ref{fig:overview_ex1} represents \textbf{Masked Sentences} generated after performing 5 operations in Figure \ref{fig:overview}. Figure \ref{fig:overview_ex2} represents \textbf{Augmented Sentences} in Figure \ref{fig:overview}. $\mathrm{<MASK>}$ means some words in the linearized sentence being masked.}
\label{fig:overflow}
\end{figure*}
\section{Method}
The overview of RoPDA is shown in Figure \ref{fig:overflow}. Firstly, We begin by preprocessing the original data into linearized sentences with entities constrained by their types. 
Secondly, we add prompt vectors to the seq2seq PLM and fine-tune it with the linearized few-shot data.
Thirdly, linearized sentences undergo strategic masking using five fundamental augmentation operations.
Fourthly, we regenerate the masked sentences using the fine-tuned PLM to produce augmented sentences.
Subsequently, Self-Consistency Filtering is applied to filter out noisy low-quality samples.
Lastly, to better utilize generated adversarial examples, a mixup is employed to interpolate the adversarial examples and the original data during the NER model training. 
\subsection{Data Preprocessing} \label{template_sec}
Similar to \citet{paolini2021structured} and \citet{zhou2022melm}, we adopt a linearization strategy to convert a sentence and its corresponding tags into a linearized sequence. As shown in Figure \ref{fig:overview}, given a sentence $X=[x_1,x_2,x_3,\cdots,x_L]$, for each entity $e_{ij}=x_i \cdots x_j$ with type $l_k$ (\emph{e.g.}, $l_k$=``PER") , we convert it into ``$[\,x_i \cdots x_j \,| \,O(l_k) \,]$", where $O(l_k)$ is the natural language form of label $l_k$ (\emph{e.g.}, $O(l_k)$=``person") . Before being sent to PLMs, labeled sentences must be processed using this template. This data preprocessing enables the PLM to explicitly take label information into account when generating tokens, thereby constraining entity and entity label consistency. Furthermore, such a template is reversible, enabling us to recover the sentence and its labels from the linearized sequence.
\subsection{Prompt-based PLM}
Fine-tuning is the prevalent way to adapt large PLMs to downstream tasks. In low-resource scenarios, however, the available training data is inadequate relative to the model size, and updating all parameters of the model can readily cause overfitting, leading to a decline in generalization.
By prepending instructions to the task input and directly generating the task output from PLMs, prompt can effectively harness the potential of PLMs, especially in low-resource settings \cite{brown2020language,liu2021gpt,liu-etal-2022-p}. Following \citet{wang2022promda}, we add a sequence of continuous trainable vectors, which is also called Soft Prompt, as shown in Figure \ref{fig:overview}, to each layer of the PLM. During training, we only update the parameters of the prompt vectors and fix all PLM parameters. 
Note that in this paper, we choose the T5 \cite{raffel2020exploring} model as our backbone sequence-to-sequence PLM.
Different from \citet{wang2022promda}, we do not pre-train prompt vectors with additional corpus texts but only utilize a small number of labeled data for training.
Although our method does not leverage external knowledge, we still achieve significant improvement compared to their method. 
\subsection{Data Augmentation} \label{subsec:3_3}
Tokens in a sentence can be divided into entity segments and context segments depending on whether or not they are in an entity. Formally, a sentence X is divided into $C_1 \,E_1 \, \cdots C_n \,E_n \,C_{n+1}$, with $C_i$ representing the context segment and $E_i$ representing the entity segment. 
To enhance data diversity, we propose five fundamental data augmentation operations, one of which is dedicated to context augmentation, while the other four focus on entity augmentation, involving label-flipping and label-preserving operations. 
The sentence is linearized according to Section \ref{template_sec} before each operation. 
Then we apply the five operations to the linearized sentence by masking the according part of it.
PLMs are then employed to generate the masked parts in order to obtain the synthetic sentences. Figure \ref{fig:overview_ex1} shows examples of each operation.
\begin{table}
\centering
\resizebox{0.9\linewidth}{!}{
\begin{tabular}{rc} 
\toprule
Strategy       & Operations                   \\ 
\hline
SA   & Op1$*$M + Op5$*$N                  \\
ELC & Op2$*$K + Op1$*$(M-K) + Op5$*$N        \\
EA  & Op3$*$K + Op1$*$(M-K) + Op5$*$N        \\
ER  & (Op3 + Op4)$*$K + Op1$*$(M-K) + Op5$*$N  \\
\bottomrule
\end{tabular}
}
\caption{Data augmentation strategies. $*$ means that the operation is repeated, followed by the number of repetitions. + means that two operations are performed sequentially. K is the number of label-flipping operations performed on each sentence. M and N represent the times of entity and context augmentation for each sentence.}
\label{aug_strategies}
\end{table}
\\
\textbf{Op1: Augmenting the Entity-Related Span} Choose an entity segment at random from the sentence, mask the entity segment and a portion of its surrounding context segments.
\\
\textbf{Op2: Changing the Entity Type} Choose an entity segment at random from the sentence and replace its type with a new entity type $l_{new}$. Then mask the entity segment and a portion of its surrounding context segments.
 \\
 \textbf{Op3: Adding an Entity} Choose an entity segment $E_i$ from the sentence randomly, and add a new entity segment of type $l_{new}$ and corresponding context segments after $E_i$.
As a result, the processed sentence would be: 
$\cdots C_i \, E_i \,\mathrm{<MASK>} \,$\\
$[\mathrm{<MASK>}\,|\, O(l_{new}) ]\,\mathrm{<MASK>}$ $ C_{i+1} \cdots$.
\\
 \textbf{Op4: Erasing an Entity} Choose an entity segment $E_i$ at random, then mask it together with a portion of its surrounding context $C_i$ and $C_{i+1}$, which is equivalent to removing the entity and associated context from the sentence. As a result, the processed sentence would be: 
 $\cdots C_{i-1} \,E_{i-1} \,\mathrm{<MASK>}\, E_{i+1} \cdots$.
 \\
 \textbf{Op5: Augmenting Contextual Spans} Choose a context segment from the sentence
at random, and then mask a portion of the context.

 Op2 to Op4 are the label-flipping operations, since they essentially alter the entity type sequence of the original sentence, thus generating an adversarial example for the altered entity type.
 There are two ways to choose the new entity type $l_{new}$ for label-flipping operations: one is to choose at random from the label set, and the other is to choose based on the entity type similarity. More details can be seen in Section \ref{subsec:ablation_flip_scheme}.
Op1, on the other hand, is a label-preserving operation that increases entity diversity by regenerating entities into new entities of the same type without altering the entity sequence.
Op5 serves to enhance the diversity of the context, a factor that we consider also crucial for increasing the overall training set diversity.

Both label-flipping and label-preserving operations can improve the performance of NER models and can be effectively utilized in combination. 
Therefore, we propose four data augmentation strategies based on these fundamental operations, as shown in Table \ref{aug_strategies}. The first, \textbf{S}tandard \textbf{A}ugmentation (SA), is the label-preserving strategy, and the latter three, \textbf{E}ntity \textbf{L}abel \textbf{C}hange (ELC), \textbf{E}ntity \textbf{A}dding (EA) and \textbf{E}ntity \textbf{R}eplacing (ER), are the label-flipping strategies due to the use of label-flipping operations.  As previously mentioned, the data generated by label-flipping strategies can be viewed as adversarial examples. 

\begin{figure}[t]
    \centering
\includegraphics[width=1.0\linewidth]{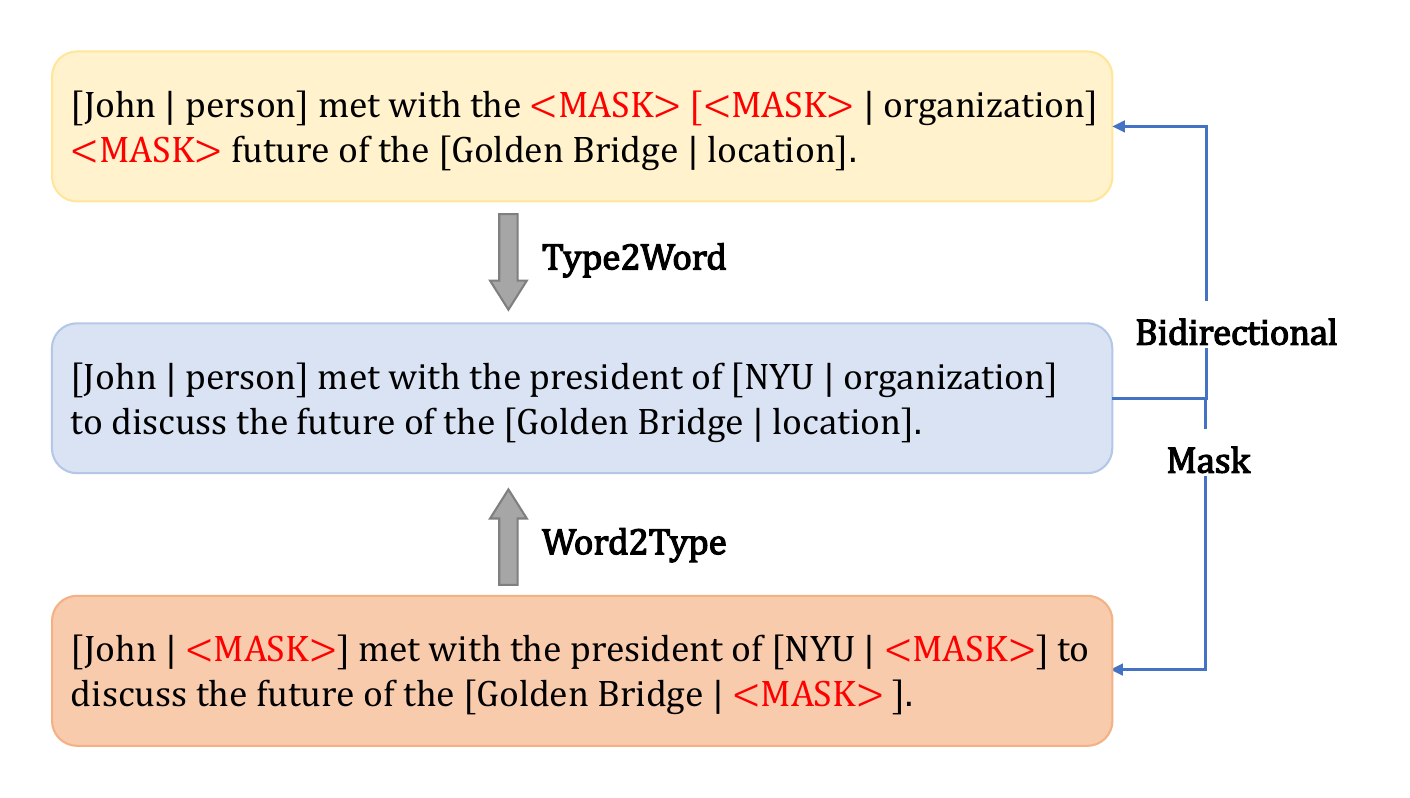}
    \caption{Bidirectional Masking}
    \label{fig:bimask}
\end{figure}
\subsection{Self-Consistency Data Filtering}
The samples generated by the proposed augmentation strategies may exhibit the entity and type inconsistency issue, especially for label-flipping strategies. 
To further improve the quality of augmented samples, we propose a novel filtering strategy based on self-consistency that employs a bidirectional mask to fine-tune the T5 model. 
As shown in Figure \ref{fig:bimask}, the bidirectional mask includes Type2Word and Word2Type, where Type2Word refers to masking the words(entities and contexts) in the linearized sentence and inferring them based on entity types, and Word2Type refers to masking entity types and inferring them based on words. After that, the fine-tuned T5 model will acquire bidirectional inference capabilities between entities/contexts and types and the ability to recognize and verify consistent samples.
Specifically, we first generate diverse augmented samples using the type-to-entity/context inference capability, as described in Section \ref{subsec:3_3}. We then use Word2Type to regenerate each entity type in the augmented samples with the assistance of the entity/context-to-type inference capability. We only keep samples whose regenerated entity types are consistent with the original. We believe that such samples are self-consistent for the model and have a higher degree of confidence.
\subsection{Mixup} \label{subsection:mixup}
Using the four strategies described in Section \ref{subsec:3_3}, we can generate a large number of adversarial samples. 
However, directly training the NER model with adversarial examples may cause overfitting on adversarial features \cite{lee2020adversarial}. 
Mixup \cite{zhang2017mixup}, as a regularization technique, can not only improve the model's generalization performance but also its robustness against adversarial attacks \cite{si2021better,lee2020adversarial}.
To this end, we leverage mixup to prevent the model from overfitting the adversarial samples and help improve generalization ability by linearly interpolating the adversarial samples and the original data. 

Given a pair of data points $(x,y)$and $(x',y')$, with $x$ denoting a data point and $y$ being its label in a one-hot representation, mixup creates a new data point by linearly interpolating the data points and their labels:
\begin{align*}
 \hat{x}&=\lambda x + (1-\lambda) x' \\
 \hat{y}&=\lambda y + (1-\lambda) y'
\end{align*}
where the mixing parameter $\lambda$ is sampled from a Beta distribution, $\lambda \sim Beta(\alpha,\beta)$.

In this work, we linearly interpolate the label-flipping sample $x_f$ with its corresponding original example $x_o$ in the hidden space. At the $m$-th layer, the hidden representations for each token in $x_f$, denoted as $h_f^m$, are interpolated with $h_o^m$, the hidden representations for each token in the original example $x_o$, by a ratio $\lambda $:
$$\hat{h}^m=\lambda h_f^m + (1-\lambda) h_o^m $$
After that, ${\hat{h}}^m$ is fed to the $(m+1)$-th layer. In the meantime, the corresponding labels of the two samples are linearly mixed with the same ratio.

\begin{table*}\small
\centering
\begin{tabular}{lcccccccccccc} 
\hline
\hline
Dataset & \multicolumn{4}{c}{CoNLL03} &  \multicolumn{4}{c}{MIT Restaurant}  &  \multicolumn{4}{c}{MIT Movie} \\
Shot & 5 & 10   & 20   & 50   & 5 & 10   & 20   & 50   & 5   & 10   & 20   & 50   \\
\hline
Baseline   & 65.9 & 73.6 & 78.5 & 82.8 & 50.3 & 59.2 & 66.1 & 70.5 & 68.0 & 70.8 & 76.0   & 81.2 \\
SDANER  & 68.7 & 74.3 & 79.4 & 83.4 & 51.2   & 59.8 & 66.2 & 70.7 & 72.8  & 75.6 & 78.1 & 81.8 \\
MELM & 67.1 & 74.6 & 78.1 & 82.9 & 50.7  & 60.1 & 66.2 & 70.4 & 69.2 & 71.3 & 76.5 & 81.5 \\
PromDA  & 71.1 & 78.2 & 81.0 & 84.2 & 51.8   & 60.3 & 66.7 & 70.7 & 75.2  & 76.4 & 78.5 & 82.4 \\
RoPDA   & \textbf{74.1} & \textbf{79.8} & \textbf{81.9} & \textbf{85.1} & \textbf{55.3}   & \textbf{62.2} & \textbf{67.6} & \textbf{71.4} & \textbf{75.5}  & \textbf{77.2} & \textbf{79.8} & \textbf{82.6} \\
\hline
MetaST\textsuperscript{*}  & 70.5 & 76.4 & 79.8 & 83.6 & 55.2   & 62.4 & 68.6 & 72.5 & 71.7  & 77.7 & 79.0   & 81.9 \\
RoPDA+self-training\textsuperscript{*} & 72.1 & 80.2 & 84.3 & 86.2 & 56.2   & 62.7 & 69.1 & 72.8 & 72.3  & 78.0   & 80.9 & 83.3 \\
RoPDA\textsuperscript{*} & \textbf{75.0}  & \textbf{81.9} & \textbf{85.2} & \textbf{86.4} & \textbf{56.6}   & \textbf{64.1} & \textbf{69.5}   & \textbf{73.0} & \textbf{75.9}  & \textbf{78.3} & \textbf{81.6} & \textbf{83.7} \\
\hline
\hline
\end{tabular}
\caption{F1-score on benchmark datasets. \textsuperscript{*} denotes the use of unlabeled data.}
\label{table:mainresult}
\end{table*}

\section{Experiments}
\subsection{Experimental Setup}

\paragraph{Datasets} 
We conduct experiments on three datasets. (a) CoNLL03 \cite{tjong-kim-sang-de-meulder-2003-introduction} is a collection of news wire articles from the Reuters Corpus, containing 4 entity types. (b) MIT Restaurant \cite{liu2013asgard} is a collection of user utterances in the restaurant domain with 8 entity types. (c) MIT Movie \cite{liu2013asgard} consists of user utterances in the movie domain with 12 entity types. These three datasets are from different domains and have varying numbers of entity types, allowing for a comprehensive evaluation of our method.
We create four low-resource settings of shot-5/10/20/50 for each dataset. In the shot-K setting, we sample K samples from the Train set for each entity type as the training set and add the remaining to the unlabeled set. 
\paragraph{Baselines} We use the model trained only on the original data as the baseline. We also adopt several other state-of-the-art methods for comparison: (1) \textbf{SDANER} \cite{dai2020analysis} explores different token replacement techniques for NER tasks. (2) \textbf{MELM\textbf} \cite{zhou2022melm} performs entity replacement on corrupted sentences based on the masked language model. (3) \textbf{PromDA} \cite{wang2022promda} employs Soft Prompt for seq2seq PLMs and proposes a dual-view data augmentation method that generates data conditioned on labels or keywords. (4) \textbf{MetaST} \cite{wang2021meta} makes use of unlabeled data through self-training and mitigates errors from noisy pseudo-labels through adaptive data re-weighting. \textbf{MetaST} is a state-of-the-art semi-supervised method.

\paragraph{Leveraging Unlabeled Data}
Unlabeled data contains a wealth of information that, if harnessed effectively, can enhance model performance significantly. 
A common solution for utilizing unlabeled data is self-training, which involves pseudo-annotating unlabeled data in the i-th iteration with the model from the (i-1)-th iteration.
To maximize the utilization of knowledge in unlabeled data, we propose a method called RoPDA\textsuperscript{*} for generating augmented samples for unlabeled data.
After pseudo-annotating unlabeled data, we only retain those with high-confidence pseudo-labels and use PLMs to synthesize augmented data for them. Finally, we train the NER model using original data, adversarial examples, pseudo-labeled data, and augmented data for unlabeled data in an iterative fashion. The overall training procedure is shown in Appendix \ref{appendix:alg}.

\subsection{Main Results}
Without unlabeled data, RoPDA achieves significant performance gains and outperforms the state-of-the-art baselines with a large margin at every shot. As shown in Table \ref{table:mainresult}, RoPDA consistently outperforms SDANER, MELM, and PromDA, and also outperforms MetaST in most cases, which uses additional unlabeled data. RoPDA yields an improvement of 2.3-8.3\% on CoNLL03, while achieving an improvement of 0.6-4.8\% on MIT Restaurant and 1.4-7.5\% on MIT Movie.

With unlabeled data, our approach achieves further improvements as shown in RoPDA\textsuperscript{*} in Table \ref{table:mainresult}.
RoPDA\textsuperscript{*} far outperforms MetaST, the state-of-the-art semi-supervised method, across all benchmarks.
In comparison to RoPDA, RoPDA\textsuperscript{*} achieves another improvement by 1.9\%, 1.7\%, and 1.1\% on three benchmarks, respectively.
To better understand the role of unlabeled data, we also conduct experiments on RoPDA+self-training.
As shown in Table \ref{table:mainresult}, RoPDA\textsuperscript{*} consistently outperforms RoPDA+self-training, which shows that data augmentation on unlabeled data can further improve model performance, reflecting the superiority of our method.
\begin{table}\small
\centering
\resizebox{0.9\linewidth}{!}{
\begin{tabular}{lccc}
\toprule
Methods  & CoNLL03 & Resta & Movie \\
\hline
RoPDA & \textbf{81.9} & \textbf{67.6}  & \textbf{79.8}  \\
 w/o. prompt & 80.5 & 67.1 & 79.0 \\
 w/o. filter & 81.5 & 67.1 & 79.5 \\
 w/o. mixup & 81.3 & 67.2 & 79.4 \\
 \bottomrule
\end{tabular}
}
\caption{Ablation study for the combination of four types under shot-20.}
\label{table:promptablation}
\end{table}

\begin{table}
\centering
\resizebox{1\linewidth}{!}{
\begin{tabular}{clcccc|c} 
\toprule
Model  & Dataset & SA & ELC & ER & EA & All \\
\hline
\multirow{4}{*}{w/o filter} & CoNLL03  & 0.2   & 0.8 & 0.4 & 0.6 & 0.4 \\
 &   \quad Resta & 0.3 & 1.0 & 0.3 & 1.5 & 0.5 \\
  &   \quad Movie & 0.2 & 0.7 & 0.6 & 0.4 & 0.3 \\
\cdashline{2-7}
  &   \quad Avg & 0.2 & 0.8 & 0.4 & 0.8 & 0.4 \\ 
  \hline
  \multirow{4}{*}{w/o mixup} & CoNLL03  & 0.2 & 1.3 & 0.5 & 0.9 & 0.6 \\
 &   \quad Resta &0.1 & 1.3 & 0.7 & 0.8 & 0.4 \\
  &   \quad Movie & 0.5 & 0.9 & 1.2 & 1.0 & 0.4 \\
  \cdashline{2-7}
  & \quad Avg & 0.3 &  1.2 & 0.8 & 0.9  & 0.5\\
\bottomrule
\end{tabular}
}
\caption{Ablation study on every single type. Values in the table represent the \textbf{drop of F1}. \textbf{Avg} represents the average of three benchmarks. \textbf{All} represents the combination of the four types}
\label{table:ablation_single}
\end{table}

\begin{table}[!ht]\small
\centering

\resizebox{0.8\linewidth}{!}{
\begin{tabular}{ccccc} 
\toprule
Dataset & SA & ELC & ER & EA  \\
\hline
CoNLL03 & 78.2 & 63.1 & 57.5 & 57.9 \\
Resta & 72.8 & 62.3 & 55.5 & 48.8 \\
Movie & 72.1 & 61.7 & 57.7 & 54.2 \\
Avg & 74.4 & 62.4 & 56.9 & 53.6 \\
\bottomrule
\end{tabular}
}
\caption{The proportion(\%) retained for each type after Self-Consistency Filtering}
\label{table:selfcon_analysis}
\end{table}
\subsection{Ablation Study} \label{sec:ablation}
As shown in Tables \ref{table:promptablation} and \ref{table:ablation_single}, we conduct ablation studies to quantify the contribution of various components, i.e., soft prompt/self-consistency filtering/mixup.
Additionally, we analyze the impact of filtering and mixup on each individual strategy.
\subsubsection{Soft Prompt}
To examine the impact of soft prompt, we compare the soft prompt model with the standard T5 model, which undergoes fine-tuning using a learning rate of 1e-4. As depicted in Table \ref{table:promptablation}, eliminating the soft prompt leads to significant performance decreases across all three benchmarks, highlighting the efficacy of soft prompt in low-resource scenarios.
\subsubsection{Self-Consistency Filtering}
After removing self-consistency filtering, as shown in Table \ref{table:promptablation}, there is a moderate decline in model performance across all benchmarks.
Subsequently, we proceed to examine the impact of self-consistency filtering on each individual strategy. 
As shown in Table \ref{table:ablation_single}, removing self-consistency filtering brings greater performance degradation to label-flipping strategies with the biggest decline in ELC having a value of 0.8, but minor degradation to the label-preserving strategy. 
We believe that this is because label-flipping strategies introduce more semantic and structural alterations to the original sentence, thereby introducing more noise and necessitating a higher degree of filtering. 
This hypothesis can be supported by Table \ref{table:selfcon_analysis}, where it is evident that self-consistency filtering retains the largest proportion of data for SA, while effectively filtering out a much greater amount of low-quality data for label-flipping strategies, specifically ELC.
\subsubsection{Mixup}
As shown in Table \ref{table:promptablation}, removing mixup leads to performance degradation across all benchmarks with an average decline of 0.5. Furthermore, an analysis of mixup's impact on each individual strategy, as shown in Table \ref{table:ablation_single}, reveals that mixup also significantly affects label-flipping strategies to a greater extent than the label-preserving strategy.
This phenomenon can be attributed to the fact that interpolating adversarial examples with the original data prevents overfitting adversarial features and improves generalization. Conversely, mixup on label-preserving data does not yield the same benefits.
Notably, self-consistency filtering and mixup have analogous impacts on these strategies, as they both essentially enhance label-flipping samples.
\begin{table}
\centering
\resizebox{1.0\linewidth}{!}{
\begin{tabular}{llll|llll} 
\toprule
SA & ELC & EA & ER & CoNLL03 & Resta & Movie & Avg \\ 
 \hline
 $\checkmark$  &  $\checkmark$   &  $\checkmark$  & & 81.0 &	66.8 &	79.5 &	75.8 \\
 $\checkmark$  &  $\checkmark$   & & $\checkmark$   &   81.3 &	67.0 &	79.4 &	75.9  \\
 $\checkmark$   &  & $\checkmark$   & $\checkmark$   & 81.6 &	\textbf{67.9} &	79.6 &	\textbf{76.4}  \\
 &  $\checkmark$   &  $\checkmark$  & $\checkmark$   &  81.7 &	67.4 &	79.6 &	76.2 \\
 \hline
  $\checkmark$  &  $\checkmark$   & $\checkmark$   & $\checkmark$   &   \textbf{81.9} &	67.6 &	\textbf{79.8} &	\textbf{76.4}  \\
\bottomrule
\end{tabular}
}
\caption{Experiment results for data combinations under shot-20.}
\label{table:combinationAblation}
\end{table}
\subsection{Analysis}
\paragraph{Different Combinations of Strategies} 
As shown in Table \ref{table:combinationAblation}, the performance of the model decreases when one strategy is removed from the combination of all, except ELC. 
This implies that the contribution of ELC is comparatively limited when incorporating the other three strategies.
We believe this is because directly changing entity types introduce too much noise, which is inferior to the two indirect ways of changing entity types, EA and ER.
In conclusion, all four strategies can improve model performance, but ELC contributes less.

\begin{table}[!ht]
\centering

\resizebox{1\linewidth}{!}{
\begin{tabular}{ccccc} 
\toprule
Dataset & F-Mixup & P-Mixup & Both-Mixup & Joint-Mixup  \\
\hline
CoNLL03 & \textbf{81.9} & 81.2 & 80.8 & 81.5 \\
Resta & \textbf{67.6} & 67.1 & 67.2 & 66.7 \\
\bottomrule
\end{tabular}
}
\caption{Different mixup methods under shot-20.}
\label{table:mixupMethodAblation}
\end{table}

\begin{table*}[!ht]
\centering
  \resizebox{\linewidth}{!}{
\begin{tabular}{lc} 
\toprule
 Strategy   & \multicolumn{1}{c}{Example}  \\ 
\hline
Original   & \textcolor[RGB]{0,196,89}{{[}Bonds \textbar{} person]} came out of Wednesday 's game against \textcolor[RGB]{0,196,89}{[New York \textbar{} organization]} in the ninth inning after suffering a mild hamstring strain .       \\
SA & \textcolor{red}{{[}Matt Carpenter \textbar{} person]} \textcolor{blue}{was carted out of the game for} \textcolor{red}{[Baltimore Orioles \textbar{} organization]} in the ninth \textcolor{blue}{with} a mild hamstring strain. \\
ELC     & \textcolor{red}{{[}Federer \textbar{} person]} \textcolor{blue}{pulled out of} Wednesday's \textcolor{red}{[European \textbar{} miscellaneous]} \textcolor{blue}{final after suffering back} strain. \\
ER    & \textcolor{red}{{[}Robinson Cano \textbar{} person]} \textcolor{blue}{left the field of }Wednesday's game \textcolor{blue}{in} \textcolor{red}{[Los Angeles \textbar{} location]} \textcolor{blue}{with two outs} in the ninth \textcolor{blue}{after a hard groundout.}       \\
EA & \textcolor{red}{{[}Marquez \textbar{} person]} \textcolor{blue}{was the} \textcolor{red}{[Boston \textbar{} organization]} \textcolor{blue}{starting pitcher and left the }game against \textcolor[RGB]{0,196,89}{[New York \textbar{} organization]} in the ninth after suffering a mild hamstring strain.  \\ 
\hline
Original   & find me a \textcolor[RGB]{0,196,89}{[nice \textbar{} rating]} place to eat that is \textcolor[RGB]{0,196,89}{[not too expensive \textbar{} price]}       \\
SA & find me a \textcolor[RGB]{0,196,89}{[nice \textbar{} rating]} place to \textcolor{blue}{have dinner} that is \textcolor{red}{[ reasonably priced \textbar{} price]}     \\
ELC     & \textcolor{blue}{where is the} \textcolor{red}{[ best \textbar{} rating]} place to \textcolor{blue}{get} \textcolor{red}{[chicken wings \textbar{} dish]}  \\
ER    & find me \textcolor{blue}{some} \textcolor{red}{[good \textbar{} rating]} \textcolor{blue}{food with} \textcolor{red}{[ parking \textbar{} amenity]} \textcolor{blue}{nearby} \\
EA & find me a \textcolor[RGB]{0,196,89}{[nice \textbar{} rating]} place to eat that is \textcolor[RGB]{0,196,89}{[not too expensive \textbar{} price]} \textcolor{blue}{and has} \textcolor{red}{[ free wifi \textbar{} amenity]} \textcolor{blue}{please}  \\
\bottomrule
\end{tabular}
}
\caption{Generated data from RoPDA. The top half of the table is from CoNLL03 and the bottom half is from MIT Restaurant. Text chunks in \textcolor[RGB]{0,196,89}{green} are the original unchanged entities. Text chunks in \textcolor{red}{red} are novel entities. Text chunks in \textcolor{blue}{blue} are novel contexts. }

\label{table:casestudy}
\end{table*}
\paragraph{Different Mixup Choices} Our original mixup method (\textbf{F-Mixup}) is to mix up the label-flipping examples, also called adversarial examples, with the corresponding original data.
We also carry out experiments on the following mixup methods to verify the superiority of F-Mixup: (1) \textbf{P-Mixup} mixups the label-preserving data with the corresponding original data. (2) \textbf{Both-Mixup} mixups both label-flipping data and label-preserving data with the corresponding original data. (3) \textbf{Joint-Mixup} mixups the label-flipping data with both the corresponding label-preserving data and original data. 
As shown in Table \ref{table:mixupMethodAblation}, \textbf{F-Mixup} is undoubtedly superior to other mixups, demonstrating the importance of interpolating adversarial examples with the original data.

\paragraph{Label Flipping Schemes}\label{subsec:ablation_flip_scheme}
There are two ways to choose $l_{new}$ for label-flipping operations. The first one \textbf{Random} is to choose randomly from the label set, and the other is to choose based on entity type similarity. For similarity-based methods, we propose two options: \textbf{Fixed} means fixed flipping of each entity type to the most/least similar type. \textbf{Probability} indicates that the flip probability is calculated based on the entity type similarity using the softmax function. 
The specific calculation method for similarity can be found in Appendix \ref{appendix:similarity_cal}.
Table \ref{table:flipAblation} shows that \textbf{Random} outperforms all similarity-based schemes.
We contend that this is due to the fact that random selection has the potential to maximize sentence diversity, whereas selection based on similarity will result in a much higher probability of flipping to a specific type than other types for each type, thus reducing sentence diversity.
\begin{table}\small
\centering

\resizebox{1.0\linewidth}{!}{
\begin{tabular}{cccccc} 
\toprule
Dataset   & Random & F+S  & P+S  & F+D  & P+D   \\
\hline
CoNLL03 & \textbf{81.9} & 81.4 & 80.8 & 80.6 & 81.2  \\
MIT       & \textbf{67.6} & 66.9 & 66.3 & 66.0 & 66.4  \\
\bottomrule
\end{tabular}
}
\caption{Ablation study for label flipping schemes. \textbf{F} is \textbf{Fixed}, \textbf{P} is \textbf{Probability}, \textbf{S} means that the more similar between two entity types, the higher the flip probability and \textbf{D} means the opposite.}
\label{table:flipAblation}
\end{table}

\begin{figure}
\centering
\includegraphics[width=1.0\linewidth]{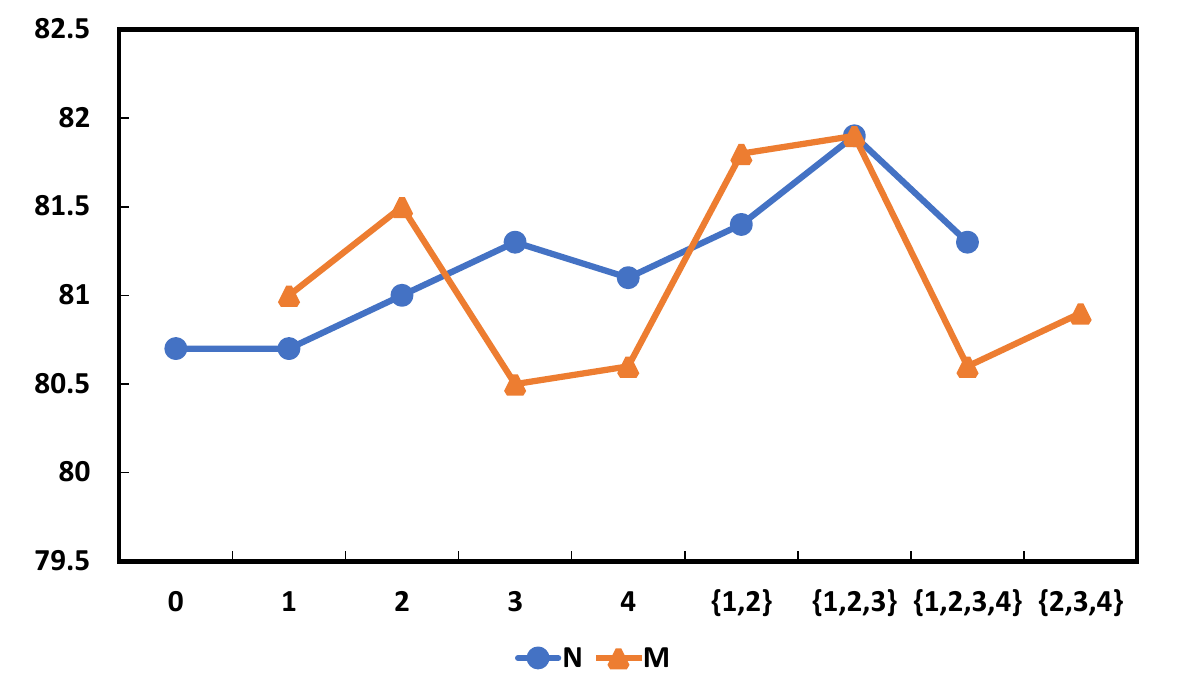}
\caption{Experiment results with different M and N on CoNLL03.}
\label{pic:MNvalue}
\end{figure}
\paragraph{Hyperparameter Setting for Data Augmentation} There are several important parameters for data augmentation: K, M and N. 
K is the number of label-flipping operations performed on each sample. K is initially set to 1, and when K increases, the F1 score on CoNLL03 falls by 1.1\%. 
M and N represent the times of entity and context augmentation for each sentence, respectively. 
As shown in Figure \ref{pic:MNvalue}, when context augmentation is removed, that is, when N=0, the performance declines by 1.2\%, indicating the importance of enhancing the contexts of sentences. 
When M and N take fixed values, the performance suffers when they are too large or too small, with M=2 and N=3 providing the best results. 
In addition, the best performance is attained  when M and N are both randomly chosen from \{1, 2, 3\}.
In our original experiment, when performing entity augmentation, we not only mask and regenerate the entity but also regenerate a portion of the surrounding context, as we believe this will make the generated entity more in tune with its context. We run experiments on CoNLL03 to verify our hypothesis. The F1 score decreases by 0.8\% when only regenerating the entity without regenerating its context, indicating the importance of regenerating its context simultaneously.
\subsection{Case Study}
Table \ref{table:casestudy} shows some generated examples from RoPDA.
RoPDA can generate high-quality new entities and increase entity diversity by leveraging the powerful inference ability and rich knowledge contained in PLMs.
Taking the sentence from CoNLL03 as an example, ``New York" is usually represented as a location-type entity, but in our example sentence, ``New York" represents an organization type. In EA strategy, the trained language model learns the true meaning of ``New York" through contextual inference and its conditioned label and thus, generates a new organization entity ``Boston" which has a similar meaning to ``New York".  
In addition, as can be seen in Table \ref{table:casestudy}, in ELC and ER strategies, the original entity can be regenerated as a new type of entity that is harmonious with the context.
Moreover, our proposed method improves not only entity diversity but also context diversity significantly.

\section{Conclusion}
In this paper, we propose a robust prompt-based data augmentation method RoPDA for low-resource NER. RoPDA performs entity augmentation and context augmentation through five fundamental augmentation operations to generate label-flipping and label-preserving examples. 
To optimize the utilization of augmented data, we introduce two techniques: Self-Consistency Filtering and mixup.
Self-Consistency Filtering efficiently eliminates low-quality samples, while mixup mitigates performance degradation arising from the direct utilization of adversarial samples. 
Extensive experiments on three benchmark datasets showcase the effectiveness of RoPDA.

\bibliography{emnlp2023}
\bibliographystyle{acl_natbib}

\appendix
\section{Training Procedure of RoPDA\textsuperscript{*}}\label{appendix:alg}
\begin{algorithm}[h]
\caption{RoPDA\textsuperscript{*} Algorithm}
\label{alg:alg1}
\textbf{Input}: few-shot labeled dataset $\mathcal{T}$; unlabeled dataset $\mathcal{U}$; the number of iterations $N$; pre-trained language model $LM$.

\textbf{Output}: a trained NER model $M$.

\begin{algorithmic}[1]
 \STATE $LM \gets \mathrm{Train}(LM, \mathcal{T})$
 \STATE $\mathcal{T}_{a0} \gets \mathrm{Augment}(LM,\mathcal{T})$
 \STATE $\mathcal{T}_{af} \gets \mathrm{SelfConsistencyFilter}(LM,\mathcal{T}_{a0})$
 \STATE $M^0 \gets \mathrm{TrainNER}(\mathcal{T}_{af})$
 \FOR{ $k \gets 1 \,\textbf{to}\, N $}
   \STATE $\hat{\mathcal{T}}_{af} \gets \mathrm{HighConfSelect}(M^{k-1},\,\mathcal{T}_{af})$
   \STATE $M^{k} \gets \mathrm{TrainNER}(\mathcal{T} \cup \hat{\mathcal{T}}_{af})$
 \ENDFOR
 \IF { $\mathcal{U} $  }
   \STATE $\mathcal{T}_u \gets \mathrm{Annotate}(M^N,\,\mathcal{U})$
   \STATE $\mathcal{T}_{au} \gets \mathrm{Augment}(LM,\,\mathcal{T}_u)$
   \FOR{ $k \gets 1 \,\textbf{to}\, N $}
 \STATE $\hat{\mathcal{T}}_A \gets \mathrm{HighConfSelect}(M^{k-1},\,\mathcal{T}_{af} \cup \mathcal{T}_{au})$
 \STATE $\mathcal{T}_u \gets \mathrm{Annotate}(M^{k-1},\,\mathcal{U})$ 
 \STATE $M^{k} \gets \mathrm{TrainNER}(\mathcal{T} \cup \hat{\mathcal{T}}_A \cup \mathcal{T}_u)$
   \ENDFOR
 \ENDIF
 \STATE $M \gets M^{N}$
 \STATE \textbf{return} $M$
\end{algorithmic}
\end{algorithm}
\section{Implementation Details}
We utilize the T5-Large \cite{raffel2020exploring} model as our generative model. The T5-Large model requires no further fine-tuning, and the prompt parameters are only fine-tuned using few-shot data. Following \cite{wang2022promda}, we use Adafactor \cite{shazeer2018adafactor} optimizer with learning rate 1e-3 and weight decay 1e-5.
We set the batch size 16 and train the model for 10k steps. When performing data augmentation, we set K to 1, and M and N are chosen at random from \{1, 2, 3\}.
We treat the NER task as a sequence labeling task and utilize the BERT-BASE model as our backbone model. When training the NER model, we set the learning rate 5e-5 and batch size 8. The hidden layer of mixup is randomly selected from\{8, 9, 10\}. The $\alpha$ and $\beta$ in the mixup are set to 130 and 5, respectively. The NER training settings of baselines are set to the same as RoPDA. We report the average micro-F1 score for 3 runs.
\begin{table}
\centering
\resizebox{1.0\linewidth}{!}{
\begin{tabular}{lccc}
\toprule
Dataset  & 1000 & 2000 & All Data \\
\hline
CoNLL03 & 87.5(+0.4) & 88.7(+0.3) & 90.3(+0.3)  \\
 Resta & 75.3(+1.6) & 78.2(+0.6) & 80.0(+0.5) \\
Movie & 83.9(+1.2) & 84.7(+0.5) & 87.9(+0.4) \\
 \bottomrule
\end{tabular}
}
\caption{Performance gains brought by RoPDA in data-rich settings.}
\label{table:data_rich}
\end{table}
\section{RoPDA in Data-Rich Settings}
Despite being designed for low-resource settings, RoPDA can still offer appreciable performance improvements in high-resource settings.
As shown in Table \ref{table:data_rich}, in data-rich scenarios, RoPDA improves model performance on all three benchmarks, and as the amount of data increases, the performance gain brought by RoPDA continues to decline.
In Addition, we discover that the worse the baseline performance of a certain dataset, the greater the performance gain brought by RoPDA.

\section{Calculation of Entity type Similarity}\label{appendix:similarity_cal}
Given two entity types, $l_1$ and $l_2$ (e.g., PER and LOC), we first obtain their natural language forms, $O(l_1)$ and $O(l_2)$ (e.g., person and location), and then employ BERT \cite{devlin-etal-2019-bert} to obtain the embedding representations, $h(l_1)$ and $h(l_2)$, for $O(l_1)$ and $O(l_2)$. Subsequently, we compute the Euclidean distance or cosine similarity between $h(l_1)$ and $h(l_2)$, considering the negative Euclidean distance or cosine similarity as the similarity measure for the entity types.

\end{document}